\documentclass[11pt]{article}

\usepackage[final]{acl}
\usepackage{xspace}
\usepackage{times}
\usepackage{latexsym}
\usepackage{amssymb}

\newcommand{\methodname}{{ProMem}\xspace}
\usepackage{tabularx}
\usepackage{tcolorbox}
 \usepackage{balance}
\usepackage{placeins}
\tcbuselibrary{breakable}
\usepackage[T1]{fontenc}
\usepackage{newfloat}
\usepackage{listings}
\usepackage{xcolor}
\usepackage{colortbl}
\usepackage{booktabs}
\usepackage{multirow}
\usepackage{amsmath}
\usepackage[utf8]{inputenc}
\usepackage{algorithm}
\usepackage{algorithmic}
\usepackage{soul}

\usepackage{microtype}

\usepackage{inconsolata}

\usepackage{graphicx}

\title{Beyond Static Summarization: Proactive Memory Extraction\\ for LLM Agents}

\author{
    Chengyuan Yang$^\dagger$ \quad 
    Zequn Sun$^{\dagger,\,}$\thanks{Corresponding author} \quad 
    Wei Wei$^\dagger$ \quad 
    Wei Hu$^{\dagger,\,\ddagger}$ \\
    $^\dagger$ State Key Laboratory for Novel Software Technology, Nanjing University, China \\
    $^\ddagger$ National Institute of Healthcare Data Science, Nanjing University, China \\
    \texttt{\{cyyang, weiw\}.nju@gmail.com, \{sunzq, whu\}@nju.edu.cn} 
}

\begin{document}
\maketitle
\begin{abstract}
Memory management is vital for LLM agents in long-term and personalized interactions.
Most previous work studies how to retrieve and use memory, but pays less attention to how memory is extracted.
We find two main limitations in existing methods.
First, extraction is ``ahead-of-time'': the agent saves information before it knows future tasks.
A single summary prompt often mixes details, events, and relations, so useful information is lost.
Second, extraction is usually one-off. Without verification, errors and hallucinations may stay in memory for a long time.
To address these limitations, we propose ProMem, a proactive memory extraction framework.
It separates details, events, and relations, and uses different extraction strategies for each type.
It also checks completeness to recover missed events and verifies facts at the atomic level to reduce hallucinations.
Experiments show that ProMem improves memory completeness and QA accuracy, while keeping a good balance between quality and token cost.
\end{abstract}

\section{Introduction}\label{sect:intro}

Memory management is a cornerstone for large language model (LLM) based agents, enabling them to maintain long-term dialogue history and personalized information of users \cite{memory_survey,luo2024dricl}.
To overcome the limitations of the LLM context window, existing research has relied on summary-based memory management \cite{Mem0}. 
These methods compress historical interactions into concise summaries, aiming to preserve essential information while reducing computational overhead.
As shown in Figure~\ref{fig:pipeline}, 
most current summary-based methods focus heavily on the memory organization and utilization phase, often designing complex memory evolution loops and advanced memory retrieval mechanisms to handle stored memory summary~\cite{MemoryOS,FlexiblyMem,LightMem}.
However, the initial memory extraction is frequently overlooked.
We argue that these methods suffer from two fundamental limitations:

\begin{figure}[t]
  \centering
  \includegraphics[width=\linewidth]{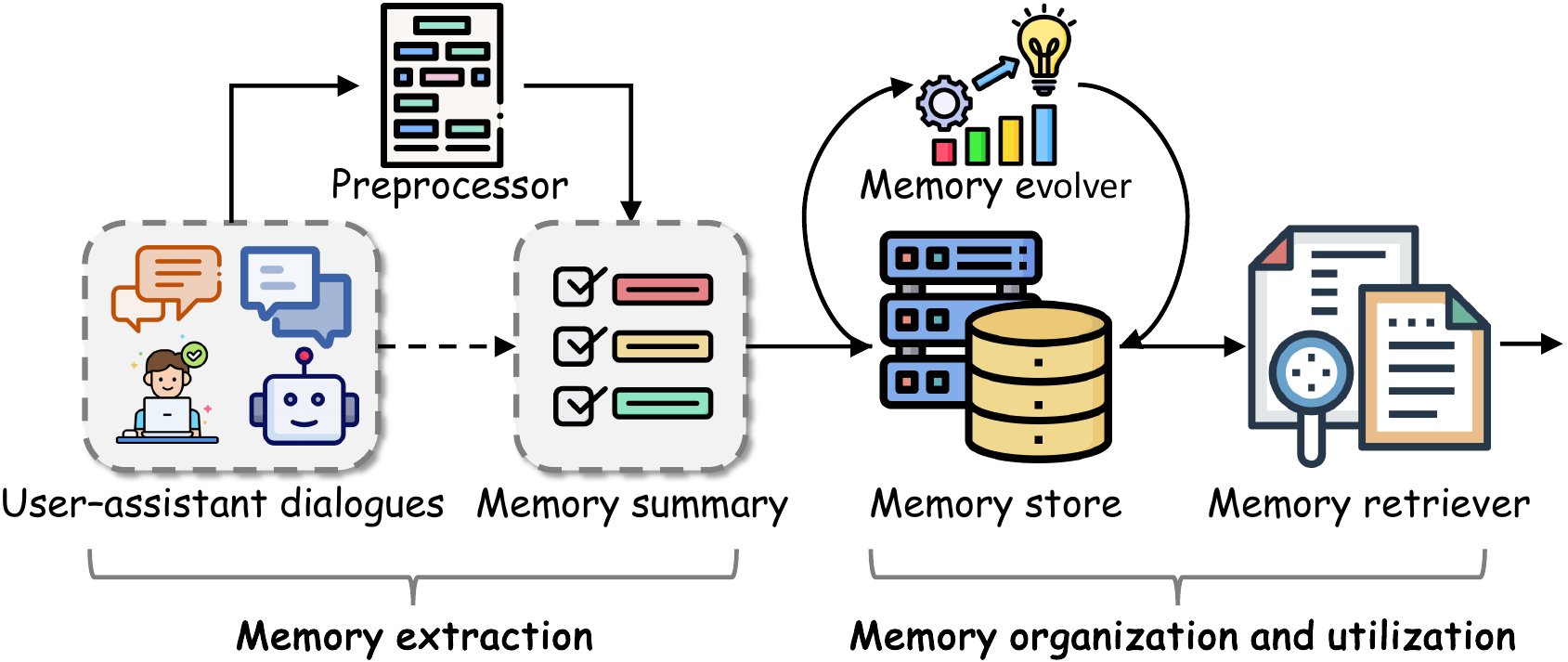}
  \caption{
  A general pipeline of summary-based memory management. It includes memory extraction, storage, update, and retrieval. Most previous work focuses on the later stages, while the extraction stage is simplified.
  }
  \label{fig:pipeline}
\end{figure}

First, memory extraction is ``\textit{ahead-of-time}''. 
The agent summarizes the dialogue history before knowing what the future task will be. 
This limitation often leads to a continuous accumulation of information loss. 
As the agent cannot predict future questions, 
it might easily discard small but crucial details or focus on irrelevant points. This makes the memory significantly less useful when the agent actually needs to invoke it later. 
Therefore, under this \textit{ahead-of-time} constraint, 
it is important to explicitly define \textit{what} the agent needs to remember. Furthermore, the extraction of distinct types of memory information cannot be treated as a single task handled by a unified extraction strategy.

Second, existing methods usually extract memory only once. 
We call this ``\textit{one-off extraction}''. 
If the agent makes a mistake or has a ``hallucination'' when it extracts memory for the first time, this error will stay in the memory and hurt the performance of agents in the future. 
We argue that feedback verification is essential for achieving high-quality memory management, yet current research has not paid enough attention to this vital step.

\begin{figure}[t]
  \centering
  \includegraphics[width=\linewidth]{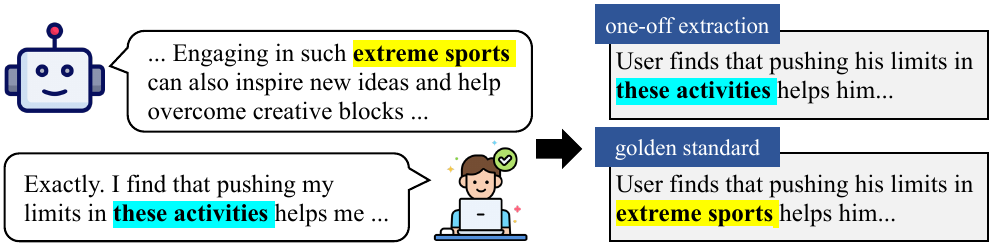}
  \caption{Example of info. loss in one-off extraction.}
  \label{fig:motivation_example}
\end{figure}

Figure~\ref{fig:motivation_example} shows a motivating example.
While the user refers to ``extreme sports'' using the phrase ``these activities,'' a single-pass extractor fails to resolve this coreference, resulting in an ambiguous memory entry compared to the golden standard.

To resolve these limitations, 
we propose \textit{proactive memory extraction}, called \methodname.
It turns memory extraction into a multi-grained and iterative process.
\methodname extracts different types of memory separately, then checks missing information and verifies suspicious facts against the original dialogue.
This process helps recover omitted content and remove hallucinations.
As a result, the final memory is more complete and more accurate.
\methodname has four parts:

\begin{itemize}
    \item \textbf{Multi-grained memory extraction}: 
    The agent scans the unstructured dialogue history to separately extract fine-grained details, core events, and their relations to form a preliminary structured memory graph.
    
    \item \textbf{Completeness check via explicit matching}: 
    The agent establishes a cross-reference matrix between the extracted events and initial memory drafts. 
    If any valid event fails to align with the draft, the agent actively recovers and integrates this overlooked information to ensure memory completeness.
    
    \item \textbf{Fact verification via atomic decomposition}:
    The agent decomposes memory entries into finer atomic facts. 
    It then utilizes natural language inference (NLI) to cross-check these facts against the source dialogue.
    It purifies invalid segments (hallucinations) without discarding the entire memory entry.
    
    \item \textbf{Relation\,verification\,and\,detail\,anchoring}:
    The agent generates probing questions to verify the logical relations (e.g., causal, temporal) between events. 
    Finally, through semantic matching, the initially uncoupled fine details are precisely re-anchored to their corresponding verified events, yielding a high-fidelity, graph-structured final memory.
\end{itemize}

To evaluate the memory extraction capability of our \methodname, we use HaluMem~\cite{HaluMem} as our main dataset. 
We chose this dataset for two main reasons. 
First, our research focuses on solving the ``missing information accumulation'' problem caused by simple one-off extraction. 
HaluMem is specifically built to detect such phenomena. 
Second, the dataset contains complex user-agent interactions that require deep understanding, which allows us to evaluate whether our multi-grained extraction and atomic-level verification can construct more accurate, high-fidelity structured memories compared to traditional methods.

Our experiments show that \methodname outperforms baselines in both memory quality and QA accuracy. It is robust to token compression and remains effective when deployed with small language models.

\section{Related Work}

\subsection{Summary-based Memory Management}
Summary-based memory management is a common way to support long-horizon tasks under limited context windows \cite{ChainRAG,ActMem,memory_survey,brain_memory_survey}.
Its goal is to compress raw interaction history into short memory texts.
Generative Agents~\cite{GenerativeAgents} uses hierarchical summarization.
MemoryBank~\cite{MemoryBank} adds a forgetting mechanism.
MemGPT~\cite{MemGPT} handles the context limit with a hierarchical virtual memory system.
Mem0~\cite{Mem0} manages user preferences and past interactions for personalized responses.
In general, these methods keep summaries in working memory and store detailed logs in long-term storage.
Many later studies focus on when to summarize and how to organize memory~\cite{MemOS,MemoryOS,Memory-R1,LightMem}.
Although these methods reduce context load, they usually treat extraction as a passive and static step.
Unlike them, \methodname focuses on making extraction more complete and reliable through proactive feedback.

\subsection{LLM-based Information Extraction}

Information extraction (IE) aims to transform unstructured text into structured formats. While traditional methods relied on supervised sequence labeling~\cite{BERT-CRF}, the emergence of LLMs has shifted the paradigm towards generative IE.

Current research largely focuses on leveraging the instruction-following capabilities of LLMs \cite{knowlm,ChatGPT4IE,OpenRelationExtraction}. 
UIE~\cite{UIE} provides a unified text-to-structure framework for different IE tasks.
CodeKGC~\cite{CodeKGC} and ChatIE~\cite{ChatIE} improve extraction with code generation or multi-turn interaction.
However, most LLM-based IE methods focus on static text, such as encyclopedic or news articles.
Our setting is different. \methodname extracts personalized user memory from multi-turn dialogue.
This task needs not only fact extraction, but also consistency and completeness checking.

\subsection{Self-Refinement and Iterative Prompting}
Recent studies show that LLMs can improve their outputs through iterative feedback.
Self-Refine \cite{Self-Refine} and Reflexion \cite{Reflexion} let the model review and revise its first output.
Similar ideas are also used in reasoning and code generation to reduce one-pass errors \cite{DynamicsPrompting,Self-Debug,Chain-of-Verification}. 
\methodname is similar in spirit, but different in goal.
Most self-refinement methods improve one task output, such as a poem or a code snippet. 
\methodname applies iteration to memory extraction in long-term dialogue.
It focuses on two problems: missing information and hallucinations in memory.

\subsection{Knowledge Graph Curation and Refinement}

Knowledge graph curation studies how to extract and refine structured knowledge through completeness checking, grounding, fact validation, entity resolution, and knowledge fusion
\cite{DBLP:journals/semweb/Paulheim17,Hogan2021,DBLP:journals/tnn/JiPCMY22,Dong2014}.
\methodname is related to this line of research but focuses on memory construction for long-term LLM agents.
\methodname applies them to different memory types at different granularities.
Details are extracted at the turn level, events are processed in topic-level contexts, and relations are verified across events.
\section{Proactive Memory Extraction}
\label{sec:3}
In this section, we first introduce our motivation,
and then present the proposed method \methodname.

\begin{figure}[!t]
  \centering
  \includegraphics[width=\linewidth]{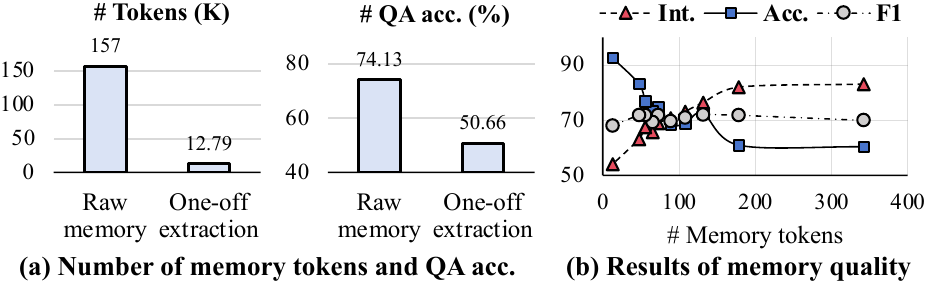}
  \caption{
  Limitations of one-off extraction. (a) It greatly reduces tokens, but QA accuracy drops by about 24\%. (b) Increasing memory size brings only limited gains, which shows a structural bottleneck.
  }
  \label{fig:motivation}
\end{figure}

\subsection{Motivation}

Current memory-augmented agents typically rely on a single-pass (one-off) extraction to compress lengthy dialogue history into concise memory. However, our empirical analysis (see Appendix~\ref{appx:setup} for setup details) on the naive RAG with raw memory and one-off extracted memory reveals two critical limitations of this pipeline:

\begin{itemize}
    \item \textbf{Information loss and accuracy drop}: 
    One-off extraction is efficient, but it loses important information.
    Figure~\ref{fig:motivation}(a) shows that it reduces memory from 157K to 12.79K tokens, but QA accuracy drops from 74.13\% to 50.66\%.
    This result shows that one scan cannot keep all necessary details.
    
    \item \textbf{Scaling\,ceiling\,of\,extracted\,memory\,quality}: 
    A simple idea is extracting more facts.
    However, Figure~\ref{fig:motivation}(b) shows that a larger memory does not bring proportional F1 gains.
    The bottleneck is not only memory size. It also comes from the one-off extraction itself.
\end{itemize}

We propose a proactive memory extraction method called \methodname.
It changes extraction from a static one-pass process to an active iterative process.
It lets the agent re-check the dialogue, recover missed details, and remove hallucinations.
This helps \methodname produce better memory with a reasonable quality-cost trade-off.

\begin{figure*}[t]
\centering
\includegraphics[width=\textwidth,height=0.26\textheight]{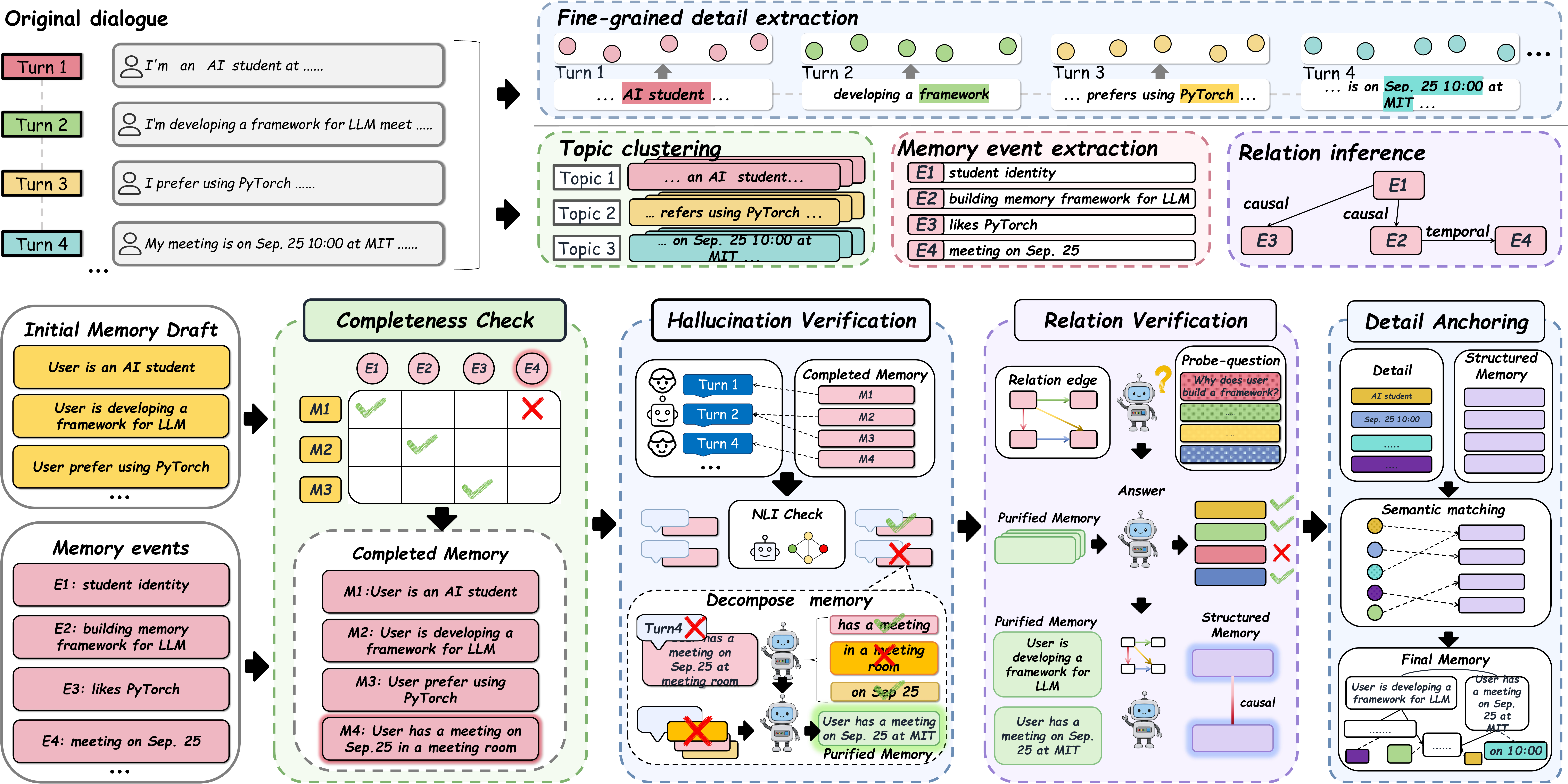}
\caption{
Overview of the \methodname framework. The pipeline consists of four main stages. 
(1) \textbf{Multi-grained Extraction}: The agent separately extracts fine details, macro events, and initial relations from the raw dialogue. 
(2) \textbf{Completeness Check}: It uses a cross-reference matrix to compare extracted events with the initial draft to recover missed information. 
(3) \textbf{Hallucination Verification}: It uses atomic decomposition and NLI to check facts and surgically remove errors. 
(4) \textbf{Relation Verification and Detail Anchoring}: It confirms logical edges and attaches fine details to the purified events. The final output is a high-quality, graph-structured memory.
}
\label{fig:framework}
\end{figure*}

\subsection{Multi-Grained Memory Disentanglement}
\label{sec:disentanglement}
The agent extracts memory at multiple granularities.
Instead of using one prompt for everything, we divide the task into three sub-tasks.
Given a dialogue history $D = \{t_1, t_2, ..., t_n\}$, our goal is to build the basic parts of a structured memory graph.

\paragraph{Fine-grained detail extraction.}
The agent scans $D$ and extracts atomic details, e.g., times, locations, and entities.
We use prompt $P_{detail}$ for this step:
\begin{equation}
    F_{detail} = \text{LLM}(D, P_{detail}),
\end{equation}
where $F_{detail}$ represents the raw set of details.

\paragraph{Topic clustering and event abstraction.}
We then cluster the dialogue turns $t_i \in D$ into topic groups $C = \{c_1, c_2, ..., c_m\}$.
For each cluster $c_j$, the agent abstracts a set of higher-level events $E = \{e_1, e_2, ..., e_k\}$.
These events describe the user's main states, actions, or stories.

\paragraph{Relation inference.}
Finally, to capture the dynamic progression of the user's memory, 
we infer logical relations, such as causal or temporal links, between events in $E$.
This gives a set of initial relation edges $R = \{(e_i, r, e_j)\}$.

\paragraph{Parallel execution.}
To reduce latency, detail extraction and event abstraction run in parallel.
Relation inference is done after event abstraction.

\subsection{Explicit Completeness Check}
\label{sec:completeness}

Many methods first produce an initial memory draft $M_{draft} = \{m_1, m_2, ..., m_p\}$.
However, this draft may miss important events.
We therefore add an explicit completeness check to compare the abstracted events $E$ with $M_{draft}$.

\paragraph{Cross-reference matrix.}
First, we map the abstracted events back to the draft.
We encode both $E$ and $M_{draft}$ with a text embedding model.
Let $v_{e_i}$ and $v_{m_j}$ be the embeddings of event $e_i$ and draft entry $m_j$.
We compute their cosine similarity:
\begin{equation}
    S(e_i, m_j) = \frac{v_{e_i} \cdot v_{m_j}}{\|v_{e_i}\| \|v_{m_j}\|}.
\end{equation}

For each event $e_i$, we find the most similar draft entry and compare the score with a threshold $\tau_{comp}$:
\begin{equation}
    \text{Match}(e_i) = \max_{m_j \in M_{draft}} S(e_i, m_j) > \tau_{comp}.
\end{equation}
If $\text{Match}(e_i)$ is true, we consider that the event has been effectively captured in the draft.

\paragraph{Missing event recovery.} 
If $\text{Match}(e_i)$ is false, then $e_i$ is a missed event.
We collect all missed events into a set $E_{miss}$.
Then we merge them with the draft to form a more complete set $M_{comp}$:
\begin{equation}
    M_{comp} = M_{draft} \cup E_{miss}.
\end{equation}

\subsection{Hallucination Verification via NLI}
\label{sec:hallucination_verification}

This module seeks to reduce hallucinations.
Although $M_{comp}$ has high recall, it may still contain false facts.
We use a cascaded verification process based on natural language inference (NLI).

\paragraph{Turn-level grounding and NLI check.}
First, we match each memory entry $m \in M_{comp}$ to its most relevant dialogue turn $t_m \in D$.
Then we use NLI to verify $m$ against $t_m$.
If the result is \textit{Entailment}, we keep the entry.
If the result is \textit{Contradiction} or \textit{Neutral}, we send it to the next stage.

\paragraph{Atomic decomposition.}
Discarding a whole entry may remove valid information.
So we decompose the problematic entry $m$ into atomic facts $A_m = \{a_1, a_2, ..., a_q\}: A_m = \text{LLM}(m, P_{decomp})$, 
where $P_{decomp}$ is the prompt for atomic decomposition. 

\paragraph{NLI verification and LLM intervention.}
We then apply NLI to each atomic fact $a_i$ against the anchored turn $t_m$.
This helps us find the incorrect part of the memory:
\begin{itemize}
    \item \textbf{Evidence verified:} 
    If the NLI result is \textit{Entailment}, we keep the atomic fact.
    
    \item \textbf{LLM correction or discard:} 
     If the result is \textit{Contradiction} or \textit{Neutral}, we send $a_i$ and $t_m$ to the LLM. The LLM acts as an adjudicator: it either revises the fact or discards it.
\end{itemize}
Finally, we reconstruct the verified facts and merge them with the entries that passed the NLI check.
This gives the purified memory $M_{purified}$.

\subsection{Relation Verification and Detail Anchoring}
\label{sec:anchoring}
This final module adds missing relations and attaches fine details back to events.

\paragraph{Necessity-driven relation verification.}
Flat memory lists often lose relations between events, e.g., causality or time order.
But connecting every pair of events is expensive and often unnecessary.
So we propose \textit{necessity-driven} verification.
For each candidate edge $(m_u, r, m_v)$ between nodes $m_u, m_v \in M_{purified}$, the agent generates a probe question $q_{rel}$.
We first test whether the agent can answer $q_{rel}$ using only the flat entries in $M_{purified}$.

\begin{itemize}
    \item 
     \textbf{Implicit relation detected:} If the agent can answer the question, the relation is already clear, so we discard the edge.
    
    \item \textbf{Explicit edge supplementation:} 
    If the agent cannot answer the question, we verify the relation against the original dialogue $D$. If the relation is supported, we add it to $R_{valid}$.
\end{itemize}

\paragraph{Detail anchoring.}
Finally, we attach the fine details $F_{detail}$ back to the correct events.
Recall that we extracted these details turn by turn in the first stage. 
Also, in the NLI verification stage, we already matched each purified event $m \in M_{purified}$ to a specific turn. 
Therefore, we can directly link them using the turn IDs. 
If a detail $f$ and an event $m$ come from the same turn, we anchor $f$ to $m$. 
This step efficiently connects fine details to high-level events without complex embedding calculations.
The final output is a graph-structured memory $G_{final} = (V, E_{edge}, A_{attr})$.
Here, $V = M_{purified}$, $E_{edge} = R_{valid}$, and $A_{attr}$ are the anchored details.

\subsection{Discussion on Computational Overhead}
\label{sect:discussion}

We acknowledge that our method uses more tokens than one-pass summarization.
This is mainly because it includes event recovery and verification.
We believe this trade-off is reasonable for four reasons:
First, memory errors are more costly than extra tokens.
If the initial memory misses key facts or contains hallucinations, later tasks will also suffer.
By spending more tokens at the start, we get cleaner and more reliable memory.
Second, memory extraction is a ``write-once, read-many'' operation.
We pay the extraction cost once, but the memory can be reused many times.
So the average cost per use is acceptable.
Third, memory extraction usually runs in the background.
Unlike response generation, it does not directly block user interaction.
So in this stage, quality is often more important than latency.
Finally, the system can still be optimized.
In practice, we do not need large models for every step.
Small language models can handle decomposition, NLI checking, and semantic matching.
We can also reduce token cost with token compression, as shown in Appendix~\ref{appx:compression}.We further report the component-level token cost, number of LLM calls, and runtime in Appendix~\ref{appx:component_analysis}.
The detailed results show that the additional overhead mainly comes from the write-time verification modules and is paid once during memory construction.

\section{Experiments}
\label{sect:exp}

\subsection{Experimental Setup}

\paragraph{Datasets.}
To comprehensively assess our framework, 
we select HaluMem~\cite{HaluMem} as our primary dataset. 
Our core contribution is to optimize the memory extraction process and mitigate hallucinations. 
HaluMem is specifically designed for this purpose. 
It provides fine-grained annotations to detect memory hallucinations and evaluate the consistency between extracted facts and the original dialogue.
This aligns with our motivation.
Besides, to demonstrate the robustness of our framework, we also conduct experiments on the widely used LongMemEval-S~\cite{LongMemEval} and LoCoMo \cite{lococmo}.

\begin{table*}[t]
    \centering
    \resizebox{\linewidth}{!}{
    \begin{tabular}{l c c c c r}
        \toprule
        \textbf{Method} & \textbf{Mem. Integrity} & \textbf{Mem. Accuracy} & \textbf{Mem. F1} & \textbf{QA Accuracy} & \textbf{\# Tokens} \\
        \midrule
        Memobase~\cite{memobase} & 14.55 & 92.24 & 25.13 & 35.33 & - \\
        Supermemory~\cite{supermemory} & 41.53 & 90.32 & 56.89 & 54.07 &  - \\
    MemOS~\cite{MemOS} & \textbf{84.81} & 86.25 & \underline{79.70} & \textbf{67.23} &  - \\
        A-Mem~\cite{A-MEM} & - & - & - & 43.02 & 36.71K \\
        MemoryOS~\cite{MemoryOS} & - & - & - & 39.24 & 61.83K \\ 
        Mem0~\cite{Mem0} & {42.91} & 86.26 & 57.31 & 53.02 & 10.02K\\
        LightMem ($r=0.4$)~\cite{LightMem} & 55.21 & \underline{93.32} & 69.37 & {56.60} & \textbf{2.40K}\\
        LightMem ($r=0.6$)~\cite{LightMem} & 57.76 & 91.01 & 70.66 & 56.79 & \underline{6.58K}\\
        LightMem ($r=0.8$)~\cite{LightMem} & 58.09 & 92.70 & 71.42 & 56.45 & \underline{9.17K}\\
        LightMem~\cite{LightMem} & 57.94 & 92.33 & 71.19 & 55.81 & 10.56K\\
        \midrule
        \methodname & \underline{81.37} & \textbf{94.68} & \textbf{87.52} & \underline{63.43} & 11.59K \\
        \bottomrule
    \end{tabular}}
    \caption{Performance (\%) comparison of memory extraction and QA on the HaluMem benchmark. The best results are highlighted in \textbf{bold}. The second best results are \underline{underlined}. ``-'' indicates that the metric is untestable, as the method does not support the corresponding evaluation.}
    \label{tab:main_results}
\end{table*}

\paragraph{Metrics.} 
We report the following metrics:
\begin{itemize}
    \item \textbf{Memory Integrity}: It measures the completeness of memory extraction. It calculates the recall of ground-truth facts, indicating how many user details are successfully extracted.
    \item \textbf{Memory Accuracy}: This metric evaluates the correctness of the extracted entries. It penalizes hallucinations and factual errors, reflecting the reliability of the memory.
    
    \item \textbf{Memory F1}: It is a comprehensive metric of memory quality, calculated as the harmonic mean of Integrity and Accuracy.
    
    \item \textbf{QA Accuracy}: It measures whether the extracted memory can effectively support the agent in answering user queries.
    
    \item \textbf{\# Tokens}: It refers to the average number of tokens consumed per session, i.e., the Total Cost of Ownership for the extraction process.
\end{itemize}

\paragraph{Baselines.} 
We select three popular open-source projects to benchmark the practical performance of memory management systems: Memobase~\cite{memobase}, Supermemory~\cite{supermemory} and MemOS~\cite{MemOS}.
We also compare our method with four leading methods in the research field:
A-Mem~\cite{A-MEM}, Mem0~\cite{Mem0}, MemoryOS~\cite{MemoryOS}, and the SOTA method LightMem~\cite{LightMem}.

\paragraph{Implementation details.}
First, we use GPT-4o-mini to process the whole session. It quickly generates the initial memory draft. 
At the same time, we use a local Llama3.1-8B model for multi-grained memory disentanglement.
The disentanglement runs in parallel with the initial extraction. 
Inside the disentanglement module, the fine-grained detail extraction also runs in parallel with the event extraction branch. For topic clustering, we follow the method in LightMem \cite{LightMem}. 
After clustering, Llama3.1-8B extracts macro events and their relations.
For completeness check, we use the Qwen3-Embedding-8B model \cite{Qwen3Embedding}. It calculates semantic similarities to find missing events. We set the completeness threshold $\tau_{comp}$ to 0.6. 
For hallucination verification, we use the \texttt{roberta-large-mnli} model to perform the NLI check. 
For relation verification, we use GPT-4o-mini to generate probe questions.
During the QA phase, 
we use Qwen3-Embedding-8B to retrieve the top-20 memory entries and append timestamps to them.
To save tokens, we use LLMLingua-2 to compress the raw dialogue. Finally, we use GPT-4o combined with human review to evaluate the correctness of the generated answers. We run all experiments on a single NVIDIA A800 GPU. 

\subsection{Main Results}
\label{sec:main_results}

Table \ref{tab:main_results} presents the main results on HaluMem.
First, our method achieves the best overall memory quality. 
It reaches the highest memory accuracy and F1. 
This proves that our framework effectively filters out hallucinations and extracts highly accurate facts.
Second, \methodname strikes the best balance between memory integrity and accuracy. 
MemOS achieves the highest memory integrity and QA accuracy. However, its memory accuracy is relatively low. 
This means MemOS uses an aggressive strategy: it extracts many facts but introduces many errors. 
Memobase has good accuracy but very low integrity, meaning it misses too much information. 
In contrast, our method achieves the highest F1 score.
We maintain the top accuracy while capturing very complete information (81.37\% integrity). 

Regarding the token cost, \methodname consumes 11.59K tokens.
This cost is very close to the baseline Mem0. 
In LightMem, $r$ is the compression ratio. $r=0.4$ means keeping 40\% of the original tokens. Although LightMem($r=0.4$) uses the fewest tokens (2.40K), its memory quality and downstream QA performance are much worse than ours. Even without compression, LightMem uses 10.56K tokens, but its results hardly improve.
Its QA accuracy stays at 55.81\%. 
At a similar token budget (11.59K vs. 10.56K), \methodname completely beats the full LightMem in all metrics. 
This proves that our high performance comes from our method, not from using more tokens. Thus, our method gets the best balance between quality and cost. 

\begin{table}[t]
    \centering\Large
    \resizebox{\linewidth}{!}{
    \begin{tabular}{l c c c c c}
        \toprule
        \textbf{Method} & \textbf{Mem. Int.} & \textbf{Mem. Acc.} & \textbf{QA Acc.} & \textbf{QA Halu.} & \textbf{QA Om.} \\
        \midrule
        \methodname (Full) & \textbf{81.37} & \textbf{94.68} & \textbf{63.43} & \textbf{17.43} & \textbf{19.14}\\
        \midrule
        \, w/o CC & 69.48 & 94.51 & 58.45 & 18.00 & 23.55\\
        \, w/o HV & 80.84 & 93.43 & 60.35 & 21.40 & 18.25\\ 
        \, w/o DA & 80.91 & 94.48 & 57.80 & 19.50 & 22.70\\
        \, w/o Rel & 81.15 & 94.42 & 61.12 & 18.40 & 20.48\\
        \, One-off Extract & 54.03 & 92.64 & 50.60 & 25.13 & 24.27\\
        \bottomrule
    \end{tabular}}
    \caption{Ablation study of key modules in \methodname.}
    \label{tab:ablation}
\end{table}

\subsection{Ablation Study}
Table~\ref{tab:ablation} shows the ablation study results.
``CC'' means Completeness Check.
``HV'' means Hallucination Verification.
``DA'' means Detail Anchoring.
``Rel'' means Relational Edges.
We also report QA Halu. and QA Om.
First, the \textit{``One-off Extract''} baseline performs worst.
It has the lowest Integrity and QA Accuracy.
Its QA Omission is also the highest.
This shows that one-pass extraction misses much information.
Second, the \textit{``w/o CC''} variant shows the value of completeness check.
Without CC, Integrity drops from 81.37\% to 69.48\%.
QA Omission also increases.
This shows that CC recovers missed events.
Third, the \textit{``w/o HV''} variant shows the value of fact checking.
Without HV, Memory Accuracy is the lowest.
QA Hallucination also rises to 21.40\%.
This shows that HV removes wrong facts.
Fourth, the \textit{``w/o DA''} shows the value of fine-grained details.
Its memory scores stay close, but QA drops a lot.
This means details are important for specific questions.
Fifth, the \textit{``w/o Rel''} variant shows the value of relation edges.
Removing relations changes memory scores only a little, but QA drops to 61.12\%.
This shows that relations help reasoning.
Finally, the full \methodname performs best.
It gives the highest QA Accuracy.
It also keeps hallucination and omission low. We also provide a component-level analysis in Appendix~\ref{appx:component_analysis}.
The results show that CC, HV, and REL detect different types of errors during memory construction.

\begin{table}[t]
    \centering
    {\small
    \begin{tabular}{l c c c}
        \toprule
        \textbf{Method} & \textbf{Memory Int.} & \textbf{Memory Acc.} & \textbf{QA Acc.} \\
        \midrule
        Mem0 & 30.59 & 82.56 & 38.41 \\
        \methodname & {43.09} & 82.32 & {49.15} \\
        \bottomrule
    \end{tabular}}
    \caption{Results using SLMs.}
    \label{tab:slm_results}
\end{table}

\begin{figure}[t]
  \centering
  \includegraphics[width=\linewidth]{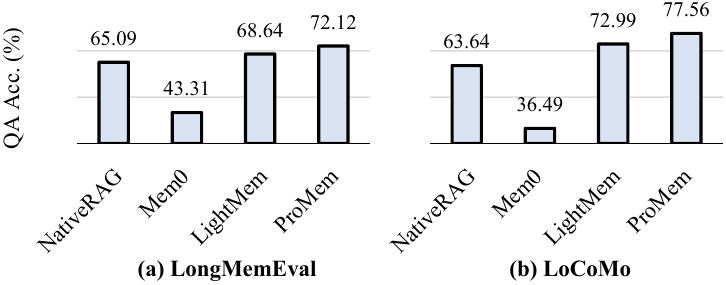}
  \caption{Results on LongMemEval-S and LoCoMo.}
  \label{fig:longmemeval}
\end{figure}

\subsection{Results with SLMs}
We test \methodname with Llama3.1-8B for extraction and QA.
We want to know whether it still works with a smaller model.
Table~\ref{tab:slm_results} shows the results.
First, \methodname is much better than Mem0 on Memory Integrity.
It can still recover missing details with a smaller model. 
Second, \methodname keeps similar Memory Accuracy.
This means the extra facts do not add many hallucinations.
Finally, QA Accuracy improves by 10.74\%.
These results show that \methodname still works well with SLMs.
\subsection{Results on LongMemEval and LoCoMo}
While HaluMem focuses on evaluating the integrity and faithfulness of extracted memory, we also need to verify the effectiveness of our method in supporting downstream applications. 
Therefore, we further conducted an additional experiment on LongMemEval-S \cite{LongMemEval} and LoCoMo \cite{lococmo}.
We compare our \methodname with NativeRAG, Mem0, and the SOTA method LightMem.
The results are presented in Figure~\ref{fig:longmemeval}.

First, \methodname achieves the best performance. 
It reaches a QA accuracy of 72.12 \% on the dataset LongMemEval-S, surpassing all baselines. 
This proves that our proactive memory extraction framework effectively captures critical information needed for answering user questions.

Second, \methodname outperforms the SOTA method LightMem. 
It is worth noting that LightMem focuses on optimizing the retrieval mechanism (how to find memory), while our work focuses on optimizing the extraction quality (what to save). 
The fact that ProMem slightly beats LightMem suggests a fundamental insight: ``\textit{Better data is more important than better algorithms}''.
By ensuring the memory is complete and accurate, we can achieve superior results even without complex retrieval tricks.

Third, the failure of Mem0 highlights the importance of extraction quality. Mem0 performs significantly worse than NativeRAG.
This indicates that Mem0's aggressive summarization likely discarded too many details, making the memory less useful than the raw chunks used by NativeRAG.
In contrast, \methodname significantly outperforms NativeRAG, demonstrating that \textit{good memory management is beneficial, while bad memory management is worse than having no memory at all}. To provide a more fine-grained comparison, we report the category-level results on LongMemEval-S and LoCoMo in Appendix~\ref{appx:category}.

\begin{table*}[!t]
    \centering
    \small
    \resizebox{\linewidth}{!}{
    \begin{tabular}{clllll}\toprule
        & \textbf{Question} & \textbf{Gold evidence} & \textbf{Gold answer} & \textbf{Extracted memory} & \textbf{Final answer} \\
        \midrule
        \multirow{2}{*}{\rotatebox[origin=c]{90}{\parbox[c]{2.3cm}{\centering Mem0}}}
        & \begin{minipage}[t]{2.5cm} 
        What role did Sarah Garcia's social network play in her career transition to Senior Manager at Oracle by Dec 10, 2025? 
        \end{minipage}
        & \begin{minipage}[t]{5cm} 
        (1) Sarah Garcia's social network expansion has been instrumental in providing insights and support during her transition to Senior Manager at Oracle. \\ 
        (2) Sarah Garcia expressed gratitude for the emotional support from friends Matthew and Linda, which has been crucial during her career transition. 
        \end{minipage}
        & \begin{minipage}[t]{1.6cm}
        Provided insights and emotional support.
        \end{minipage}
        & \begin{minipage}[t]{4cm} 
        Sarah Garcia is transitioning to a Senior Manager position at Oracle.
        \end{minipage}
        & \begin{minipage}[t]{1.3cm} \textcolor{red}{Positive influence on career growth.}. \end{minipage} \\
        
        \midrule
        \multirow{2}{*}{\rotatebox[origin=c]{90}{\parbox[c]{2.3cm}{\centering \methodname}}}
        & \begin{minipage}[t]{2.5cm} 
        What role did Sarah Garcia's social network play in her career transition to Senior Manager at Oracle by Dec 10, 2025? 
        \end{minipage}
        & \begin{minipage}[t]{5cm} 
        (1) Sarah Garcia's social network expansion has been instrumental in providing insights and support during her transition to Senior Manager at Oracle. \\ 
        (2) Sarah Garcia expressed gratitude for the emotional support from friends Matthew and Linda, which has been crucial during her career transition. 
        \end{minipage}
        & \begin{minipage}[t]{1.6cm}
        Provided insights and emotional support.
        \end{minipage}
        & \begin{minipage}[t]{4cm} 
        \textbf{Event}: Transition to Senior Manager/Support from friends\\
        \textbf{Details}: Friends Matthew and Linda/Dec 10, 2025/providing insights and support/at Oracle\\
        \textbf{Relation}: [Support from friends] $\xrightarrow{}$ [Transition to Senior Manager].
        \end{minipage}
        & \begin{minipage}[t]{1.3cm} \textcolor{blue}{Provided diverse perspectives and support}. 
        \end{minipage} \\
        \bottomrule
    \end{tabular}
    }
    \caption{A case study comparing the memory extraction and QA performance of Mem0 and \methodname.}
    \label{tab:examples}
\end{table*}

\subsection{ProMem as a Memory Foundation}

\begin{figure}[t]
    \centering
    \includegraphics[width=\linewidth]{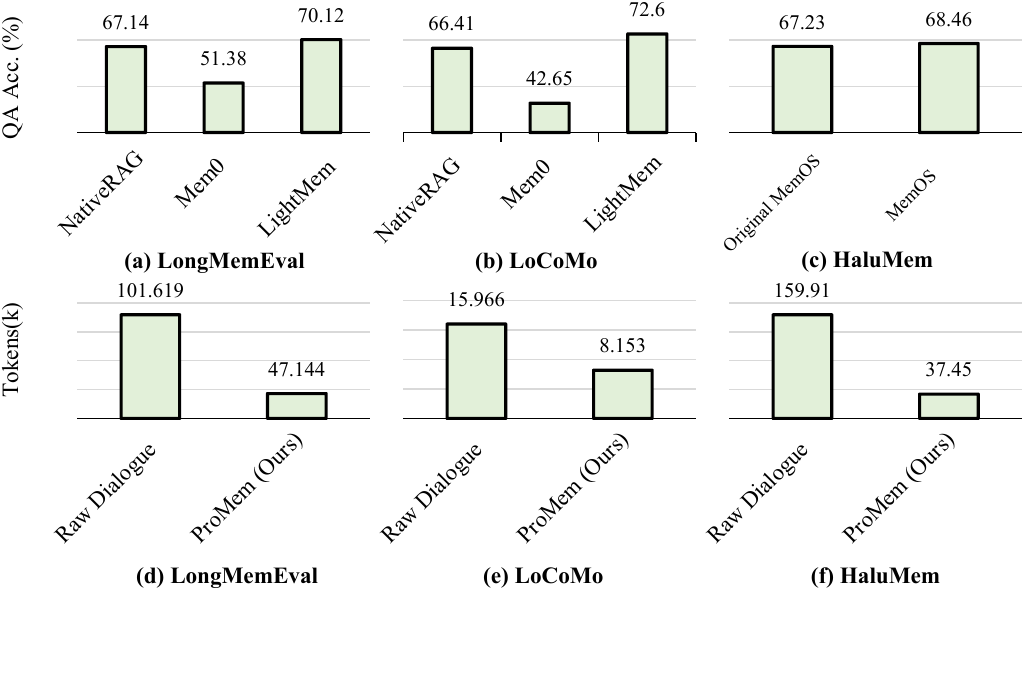} 
    \caption{(a) and (b) show the QA accuracy when replacing the raw dialogue with our extracted memory on LongMemEval-S and LoCoMo.
(c) compares the original MemOS with MemOS using ProMem memory on HaluMem.
(d)--(f) show the reduction in input tokens on the three datasets.
Overall, ProMem provides a compact and high-quality memory foundation for downstream memory systems.}
    \label{fig:memory_foundation}
\end{figure}

We conduct an experiment to show that ProMem is a high-quality memory foundation.
We use our extracted memory to replace the raw dialogue in existing baselines (NativeRAG, Mem0, and LightMem).
We also replace the original memory of MemOS with ProMem memory while keeping its retrieval and management pipeline unchanged.
Figure~\ref{fig:memory_foundation} shows the results.

First, \methodname substantially reduces the input size.
As shown in Figure~\ref{fig:memory_foundation}(d)--(f), the input tokens are reduced on all three datasets.
For example, on LongMemEval-S, the input size drops from 101.619K tokens for the raw dialogue to 47.144K tokens with ProMem memory.
On HaluMem, it drops from 159.91K to 37.45K tokens.
This shows that \methodname can provide a much more compact memory representation.

Second, the QA accuracy remains high.
Figure~\ref{fig:memory_foundation}(a) and (b) show that the baselines still perform well when using our extracted memory.
For example, LightMem achieves an accuracy of 70.12\% on LongMemEval-S and 72.60\% on LoCoMo.
Moreover, Figure~\ref{fig:memory_foundation}(c) shows that replacing the original MemOS memory with ProMem memory improves its QA accuracy on HaluMem from 67.23\% to 68.46\%, while keeping the retrieval and management pipeline unchanged.
This controlled comparison further demonstrates the quality of the memory extracted by \methodname.

Overall, these results show that \methodname preserves useful information while substantially reducing the input length.
It can also serve as a stronger memory input for an existing memory system.
Therefore, \methodname provides an efficient and high-quality memory foundation for downstream LLM agents.

\subsection{Case Study}
To intuitively show the advantage of our method, we present a case study in Table~\ref{tab:examples}. The question asks how Sarah Garcia's social network helped her career transition.
Mem0 fails because its extraction is too coarse.
It keeps only the general transition and misses the specific help from the network.
So its answer is vague: ``Positive influence''.
This shows that traditional summarization often drops important details.
In contrast, \methodname keeps fine-grained details.
Its recurrent verification loop extracts key evidence such as ``diverse perspectives'' and ``emotional support''.
This more complete memory helps \methodname give a precise answer that matches the ground truth.
\label{sec:case}

\section{Conclusion and Future Work}

In this paper, we identify the limitations of existing ``one-off'' memory summarization and propose proactive memory extraction. 
Our method introduces a recurrent feedback loop that enables agents to actively generate probe questions and verify facts against the dialogue. This helps the agent correct memory errors and supplement missing entries.
Extensive experiments demonstrate that our \methodname achieves SOTA performance in both memory quality and QA tasks. 
\methodname can also be efficiently deployed using SLMs.
In future work, we plan to extend our work to lifelong memory management. We will investigate memory updating and forgetting methods to manage memory.

\section*{Limitations}

Although our \methodname method demonstrates superior performance in memory extraction and downstream QA tasks, we acknowledge several limitations that need to be addressed in future research.
The first is computational overhead and latency. As discussed in Section~\ref{sect:discussion}, our method introduces a recurrent feedback loop involving self-questioning and supplementary extraction. Compared to traditional one-pass summarization, this iterative process inevitably increases token consumption and inference latency. While our analysis shows that the high quality of memory justifies this cost, this overhead might still be a bottleneck for strictly real-time or resource-constrained applications.
The second is dependence on backbone LLM capabilities. The effectiveness of our ``self-questioning'' and ``verification'' modules relies heavily on the reasoning capabilities of the backbone LLM. If the model is too weak (e.g., unable to generate high-quality probing questions), the feedback loop may fail to correct errors. Although our experiments show that Llama3.1-8B performs reasonably well, exploring the lower bound of model size for this framework remains an open question.

\section*{Acknowledgments}
This work was supported by National Natural Science Foundation of China (Nos.~62406136 and 62676187) and Natural Science Foundation of Jiangsu Province (No. BK20241246).

\bibliography{custom}

\appendix

\section{Setup for Empirical Study}\label{appx:setup}
To investigate the performance bottlenecks of traditional methods, we conduct a series of empirical experiments on the HaluMem dataset \cite{HaluMem}. 
While conventional memory management typically performs extraction at the session level (where all dialogue turns are combined into a single input), we systematically vary the extraction granularity to observe its impact. 
Specifically, we evaluate extraction intervals ranging from every 2 turns to 4, 6, and 8 turns, eventually reaching the entire session (the standard baseline). 
To isolate the impact of granularity, the model strictly operates in a one-off extraction mode across all settings, without any feedback or iterative verification.

Regarding evaluation, we adopt the following metrics as in the main experiments: Memory Integrity (Int.) represents completeness, while Memory Accuracy (Acc.) reflects the factual correctness of the extracted content. We also report the F1 score, the harmonic mean of Integrity and Accuracy, to provide a comprehensive assessment of the overall memory quality.
\section{Component-level Analysis}
\label{appx:component_analysis}

We further analyze \methodname from two aspects: the problems detected by different verification modules and their token costs.

\paragraph{Detected errors.}
Table~\ref{tab:component_error} reports the errors detected by Hallucination Verification (HV), Completeness Check (CC), and Relation Verification (REL).

\begin{table}[t]
\centering
\small
\resizebox{\columnwidth}{!}{
\begin{tabular}{lccc}
\toprule
\textbf{Model}
& \shortstack{\textbf{Hallucination}\\\textbf{by HV}}
& \shortstack{\textbf{Missing Events}\\\textbf{by CC}}
& \shortstack{\textbf{Missing Relations}\\\textbf{by REL}} \\
\midrule
GPT-4o-mini
& 108 / 1,284 (8.41\%)
& 80 / 682 (11.73\%)
& 12 / 472 (2.54\%) \\
Llama3.1-8B
& 348 / 2,199 (15.83\%)
& 114 / 682 (16.71\%)
& 17 / 472 (3.60\%) \\
\bottomrule
\end{tabular}
}
\caption{
Component-level errors detected by different verification modules.
HV, CC, and REL denote Hallucination Verification, Completeness Check, and Relation Verification, respectively.
}
\label{tab:component_error}
\end{table}

The results show that the extraction process contains different types of errors.
Both models produce unsupported facts, missing events, and missing relations.
This motivates the use of dedicated verification modules for different error types.

\paragraph{Component-level cost.}
Table~\ref{tab:component_cost} reports the average token cost of each component.

\begin{table*}[t]
\centering
\small
\setlength{\tabcolsep}{5pt}
\begin{tabular}{lcccccc}
\toprule
\textbf{Model}
& \textbf{Initial Draft}
& \textbf{CC}
& \textbf{HV}
& \textbf{DA}
& \textbf{REL}
& \textbf{Total} \\
\midrule
GPT-4o-mini
& 3.118K (26.90\%)
& 1.747K (15.08\%)
& 1.503K (12.97\%)
& 3.055K (26.36\%)
& 2.165K (18.68\%)
& 11.59K \\
Llama3.1-8B
& 3.691K (27.40\%)
& 1.822K (13.53\%)
& 2.050K (15.22\%)
& 3.330K (24.72\%)
& 2.570K (19.08\%)
& 13.47K \\
\bottomrule
\end{tabular}
\caption{
Average token cost per session for different components of \methodname.
CC, HV, DA, and REL denote Completeness Check, Hallucination Verification, Detail Anchoring, and Relation Verification, respectively.
}
\label{tab:component_cost}
\end{table*}

Among the verification modules, DA has the largest token cost because it links fine-grained details back to the corresponding events.
CC and REL also introduce additional cost for recovering missing events and verifying relations.
For the GPT-4o-mini setting, \methodname requires 4.26 LLM calls and 21.38 seconds per session on average.
These costs are incurred at write time, while the resulting memory can be reused for later queries.
\section{Results with Token Compression}\label{appx:compression}

\begin{figure}[t]
  \centering
  \includegraphics[width=\linewidth]{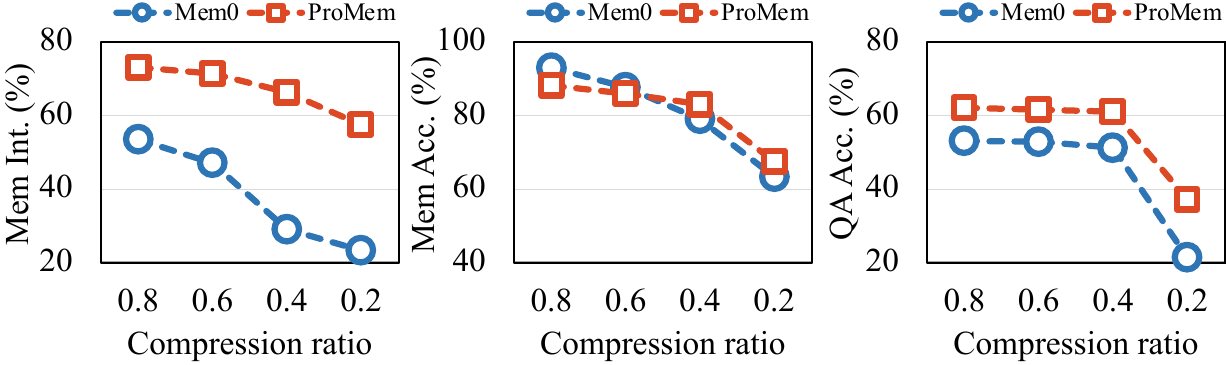}
  \caption{Performance w.r.t. compression ratios.}
  \label{fig:ratio}
\end{figure}

As discussed in Section~\ref{sect:discussion}, our extraction method inevitably introduces higher token overhead compared to one-pass methods. 
To address this efficiency concern and explore the practical deployability of \methodname, we conducted a token compression experiment.
We discarded unimportant tokens from the dialogue history at different compression ratios (0.8, 0.6, 0.4, 0.2) and compared the performance of \methodname against the baseline Mem0. 
We follow LightMem and use LLMLingua-2~\cite{LLMLingua} as the compression model. 
The results are shown in Figure~\ref{fig:ratio}.

First, our \methodname demonstrates exceptional robustness to information loss. 
As the compression ratio drops from 0.8 to 0.2 (meaning 80\% of the dialogue tokens are discarded), our method maintains a remarkably stable performance. 
Specifically, at a low compression ratio of 0.2, \methodname still achieves a QA score of 37.20\% and memory integrity of 57.20\%.
This indicates that our self-questioning and supplementary extraction can successfully extract high-quality memory even from highly compressed inputs.

Second, \methodname significantly outperforms the baseline in low-resource settings. In contrast, Mem0 is highly sensitive to token reduction. As shown in the Figure~\ref{fig:ratio}, when the ratio drops to 0.2, the performance of Mem0 collapses. 
Its memory integrity drops to 23.28\% and QA to 21.34\%.

These results strongly imply that 
\textit{we can deploy \methodname with a high compression rate in real-world applications.} 
By aggressively compressing the input tokens, we can reduce the token consumption by 60\% while still achieving superior performance compared to full-input baselines. This effectively resolves the concern regarding computational costs.

\begin{figure}[t]
  \centering
  \includegraphics[width=0.9\linewidth]{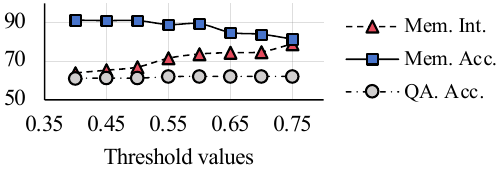}
  \caption{Performance w.r.t. different values of $\tau_{comp}$.}
  \label{fig:threshold}
\end{figure}

\section{Sensitivity Analysis on $\tau_{comp}$} \label{appx:hyperparameter}

To further evaluate the robustness of \methodname, we conduct a sensitivity analysis on the threshold $\tau_{comp}$. 
Figure~\ref{fig:threshold} shows the variations in Memory Integrity, Memory Accuracy, and downstream QA Accuracy. We observe that increasing $\tau_{comp}$ effectively enhances the completeness (Integrity) of the extracted memory by promoting supplementary extractions.
Although this leads to a slight decrease in precision (Accuracy), the downstream QA performance stays consistent, fluctuating only marginally. 
This stability demonstrates that the proactive verification loop in \methodname can effectively manage the information flow even with different threshold settings. The results justify our choice of $\tau_{comp}=0.6$, which serves as a sweet spot for maintaining high-quality memory without excessive token overhead.

\section{Category-level Results on LongMemEval-S and LoCoMo}
\label{appx:category}

\begin{table*}[!t]
\centering
\small

\setlength{\tabcolsep}{3.5pt}
\resizebox{\textwidth}{!}{
\begin{tabular}{lccccccc}
\toprule
\textbf{Method}
& \shortstack{\textbf{Temporal}}
& \shortstack{\textbf{Multi-Session}}
& \shortstack{\textbf{Knowledge-Update}}
& \shortstack{\textbf{Single-User}}
& \shortstack{\textbf{Single-Assistant}}
& \shortstack{\textbf{Single-Preference}}
& \textbf{Overall} \\
\midrule

Naive RAG
& $39.85 \pm 0.28$
& $48.48 \pm 0.32$
& $67.95 \pm 0.90$
& $90.00 \pm 0.39$
& $98.21 \pm 1.11$
& $53.33 \pm 0.21$
& $65.09 \pm 0.49$ \\

Mem0
& $29.95 \pm 0.42$
& $36.01 \pm 0.41$
& $59.92 \pm 0.32$
& $71.23 \pm 0.45$
& $30.86 \pm 0.57$
& $49.81 \pm 0.41$
& $43.31 \pm 0.42$ \\

LightMem
& $67.18 \pm 0.10$
& $71.74 \pm 0.25$
& $83.12 \pm 0.47$
& $87.14 \pm 0.37$
& $32.14 \pm 0.32$
& $68.18 \pm 0.11$
& $68.64 \pm 0.26$ \\

\methodname
& $69.43 \pm 0.20$
& $73.51 \pm 0.22$
& $83.10 \pm 2.11$
& $92.85 \pm 0.24$
& $42.67 \pm 0.08$
& $76.51 \pm 0.14$
& $72.12 \pm 0.49$ \\

\bottomrule
\end{tabular}
}

\caption{
Category-level QA accuracy (\%) on LongMemEval-S.
The best result in each column is shown in bold, and the second-best result is underlined.
}
\label{tab:longmemeval_category}

\vspace{8pt}

\setlength{\tabcolsep}{3pt}

\begin{tabular}{lccccc}
\toprule
\textbf{Method}
& \shortstack{\textbf{Single-Hop}}
& \shortstack{\textbf{Temporal}}
& \shortstack{\textbf{Open-Domain}}
& \shortstack{\textbf{Multi-Hop}}
& \textbf{Overall} \\
\midrule

Naive RAG
& $70.99 \pm 0.43$
& $56.39 \pm 0.70$
& $47.92 \pm 0.70$
& $55.32 \pm 0.73$
& $63.64 \pm 0.56$ \\

Mem0
& $38.41 \pm 0.26$
& $37.07 \pm 0.35$
& $34.38 \pm 0.73$
& $30.85 \pm 0.67$
& $36.49 \pm 0.38$ \\

LightMem
& $76.68 \pm 0.49$
& $70.58 \pm 0.53$
& $66.70 \pm 0.65$
& $66.87 \pm 0.28$
& $72.99 \pm 0.47$ \\

\methodname
& $80.50 \pm 0.20$
& $74.18 \pm 0.12$
& $69.51 \pm 0.28$
& $75.38 \pm 0.10$
& $77.56 \pm 0.17$ \\

\bottomrule
\end{tabular}

\caption{
Category-level QA accuracy (\%) on LoCoMo.
The best result in each column is shown in bold, and the second-best result is underlined.
}
\label{tab:locomo_category}

\end{table*}
Tables~\ref{tab:longmemeval_category} and~\ref{tab:locomo_category} report the category-level QA results.
On LongMemEval-S, \methodname outperforms LightMem in five of the six categories and achieves similar performance on Knowledge-Update.
On LoCoMo, \methodname achieves the best performance in all four categories.
These results show that the improvement of \methodname is consistent across different question types.
\section{Prompt Details}\label{appx:prompt}
\raggedbottom
To facilitate reproducibility, we present the prompt templates used in our \methodname framework.

\paragraph{Stage 1: Multi-Grained Disentanglement.}
The prompts for fine-grained detail extraction, event abstraction, and relation inference are shown in Figures~\ref{fig:prompt_detail}--\ref{fig:prompt_relation}.

\paragraph{Stage 2: Completeness Check.}
The prompt for initial memory draft extraction is shown in Figure~\ref{fig:prompt_draft}.

\paragraph{Stage 3: Hallucination Verification.}
The prompts for atomic decomposition and LLM intervention are shown in Figures~\ref{fig:prompt_decomp} and~\ref{fig:prompt_intervention}.

\paragraph{Stage 4: Relation Verification.}
The prompt for relation probing is shown in Figure~\ref{fig:prompt_probe}.

Note that the placeholders enclosed in curly braces
(e.g., ``\{user\_name\}'' and ``\{memory\_entry\}'')
are replaced by actual data during inference.


\noindent
\begin{minipage}{\columnwidth}
\begin{tcolorbox}[
    colback=white,
    colframe=black,
    boxrule=0.5pt,
    arc=2pt,
    left=4pt,
    right=4pt,
    top=4pt,
    bottom=4pt
]
\footnotesize

You are a micro-detail extractor.

\textbf{Dialogue Turn}: \{current\_turn\}

\textbf{Your task}: Extract all fine-grained, atomic details from this specific dialogue turn.
Focus heavily on specific entities, times, locations, and explicit object names.

\textbf{Rules}:

- Do NOT summarize the overarching event. Only extract the specific attribute fragments.

- Examples of good details: ``10:00 AM'', ``at MIT'', ``prefers PyTorch'', ``AI student''.

- Output as a simple comma-separated list of short phrases.

\textbf{Output format}: [``detail 1'', ``detail 2'', ...]
\end{tcolorbox}

\captionof{figure}{Prompt for Fine-grained Detail Extraction ($P_{detail}$).}
\label{fig:prompt_detail}
\end{minipage}

\vspace{3pt}


\noindent
\begin{minipage}{\columnwidth}
\begin{tcolorbox}[
    colback=white,
    colframe=black,
    boxrule=0.5pt,
    arc=2pt,
    left=4pt,
    right=4pt,
    top=4pt,
    bottom=4pt
]
\footnotesize

You are an overarching event summarizer.

We have grouped several dialogue turns that share the same topic.

\textbf{Topic-clustered Dialogue}: \{clustered\_turns\}

\textbf{Your task}: Abstract the core ``Memory Events'' from this clustered dialogue.
A memory event should represent a macro-level action, status, or narrative of the user
(``\{user\_name\}'').

\textbf{Rules}:

- Ignore fragmented times or locations unless they define the event.
Focus on the ``Big Picture''.

- Write each event as a clear, standalone sentence in the third person.

\textbf{Output format}: Just a list of macro events, one per line.
\end{tcolorbox}

\captionof{figure}{Prompt for Topic-based Event Abstraction.}
\label{fig:prompt_event}
\end{minipage}

\vspace{3pt}


\noindent
\begin{minipage}{\columnwidth}
\begin{tcolorbox}[
    colback=white,
    colframe=black,
    boxrule=0.5pt,
    arc=2pt,
    left=4pt,
    right=4pt,
    top=4pt,
    bottom=4pt
]
\footnotesize

You are a structural logic analyzer.

Below is a list of macro memory events extracted from a user's dialogue history.

\textbf{Extracted Events}: \{event\_list\}

\textbf{Your task}: Analyze the logical relationships between these events.
Identify any explicit or highly probable ``Causal'' (A leads to B) or
``Temporal'' (A happens before B) connections.

\textbf{Rules}:

- Only infer relationships that are strongly supported by common sense or the event context.

- If two events are unrelated, do not force a connection.

\textbf{Output format}: Output strictly as a JSON list of arrays:

[ [``Event\_A\_ID'', ``relation\_type'', ``Event\_B\_ID''], ... ]
\end{tcolorbox}

\captionof{figure}{Prompt for Relation Edge Inference.}
\label{fig:prompt_relation}
\end{minipage}

\vspace{3pt}


\noindent
\begin{minipage}{\columnwidth}
\begin{tcolorbox}[
    colback=white,
    colframe=black,
    boxrule=0.5pt,
    arc=2pt,
    left=4pt,
    right=4pt,
    top=4pt,
    bottom=4pt
]
\footnotesize

You are a Personal Fact Extraction Engine.

Your task is to analyze conversation logs to identify and catalog every piece of factual information regarding the user
(``\{user\_name\}'').
The goal is to convert dialogue fragments into explicit, self-contained statements to form an initial memory draft.

\textbf{Important instructions}:

1. Process the user messages and decide whether they contain factual information.

- Extract ANY assertion of fact, action, state, or preference.

- If no useful information is provided (e.g., pure greeting or filler), skip it.

2. Perform light contextual completion so that each fact is a clear standalone statement.

Example:

``User: Went to the gym.'' $\rightarrow$ ``User went to gym.''

3. \textbf{CRITICAL RULE}: Always use the name ``\{user\_name\}'' as the subject instead of
``User'', ``He'', ``She'', or ``I''.

\textbf{Reminder}: Be exhaustive. Output the extracted facts as a clear list.
\end{tcolorbox}

\captionof{figure}{Prompt for extracting the Initial Memory Draft.}
\label{fig:prompt_draft}
\end{minipage}

\vspace{3pt}


\noindent
\begin{minipage}{\columnwidth}
\begin{tcolorbox}[
    colback=white,
    colframe=black,
    boxrule=0.5pt,
    arc=2pt,
    left=4pt,
    right=4pt,
    top=4pt,
    bottom=4pt
]
\footnotesize

You are an expert linguist and data parser.

\textbf{Current Memory Entry}: \{memory\_entry\}

\textbf{Your task}: Decompose the complex memory entry above into a list of independent,
fine-grained ``atomic facts''.

\textbf{Rules}:

- Each atomic fact MUST contain only ONE single piece of information
(e.g., one action, one time, or one location).

- Each atomic fact must be a complete sentence with a clear subject
(``\{user\_name\}'').

- Do not add any external information or guess meanings.
Just split the existing sentence.

\textbf{Example}:

Entry: ``Sarah has a meeting on Sep. 25 at the Oracle office.''

Output:

1. Sarah has a meeting.

2. The meeting is on Sep. 25.

3. The meeting is at the Oracle office.

\textbf{Output format}: Just a list of atomic facts, one per line. No other text.
\end{tcolorbox}

\captionof{figure}{Prompt for Atomic Decomposition ($P_{decomp}$).}
\label{fig:prompt_decomp}
\end{minipage}

\vspace{3pt}


\noindent
\begin{minipage}{\columnwidth}
\begin{tcolorbox}[
    colback=white,
    colframe=black,
    boxrule=0.5pt,
    arc=2pt,
    left=4pt,
    right=4pt,
    top=4pt,
    bottom=4pt
]
\footnotesize

You are a strict factual adjudicator.

An extracted atomic fact has been flagged as potentially hallucinated or inaccurate by an NLI system.

\textbf{Source Dialogue Turn}: \{source\_turn\}

\textbf{Flagged Atomic Fact}: \{atomic\_fact\}

\textbf{Your task}: Read the source dialogue turn carefully and evaluate the flagged fact.
Choose one of two actions:

1. \textbf{REVISE}: If the fact contains a minor error
(e.g., wrong date or slight entity mismatch) but the core intent exists in the dialogue,
revise it to perfectly match the dialogue.

2. \textbf{DISCARD}: If the fact is a pure hallucination or unsupported inference
not mentioned in the dialogue, discard it entirely.

\textbf{Output format}: Output strictly as a JSON object:

\{\{``action'': ``REVISE'' or ``DISCARD'',
``revised\_fact'': ``your revised text or null''\}\}
\end{tcolorbox}

\captionof{figure}{Prompt for LLM Intervention and Correction.}
\label{fig:prompt_intervention}
\end{minipage}

\vspace{3pt}


\noindent
\begin{minipage}{\columnwidth}
\begin{tcolorbox}[
    colback=white,
    colframe=black,
    boxrule=0.5pt,
    arc=2pt,
    left=4pt,
    right=4pt,
    top=4pt,
    bottom=4pt
]
\footnotesize

You are an analytical investigator building a memory graph.

We have identified two memory events and a potential logical relation between them.

\textbf{Event A}: \{event\_A\}

\textbf{Event B}: \{event\_B\}

\textbf{Inferred Relation}: \{relation\_type\} (from Event A to Event B)

\textbf{Your task}: Generate a specific probing question to verify if this relation truly exists in the user's history.

\textbf{Rules}:

- The question must focus on the LINK between the two events.

- Example: If Event A is ``User likes Python'' and Event B is ``User learns PyTorch'',
with relation ``Causal'', the question should be:
``Why did the user choose to learn PyTorch?''
or
``Is the user learning PyTorch because they like Python?''

\textbf{Output format}: Just output the generated question. No other text.
\end{tcolorbox}

\captionof{figure}{Prompt for generating Relation Probe Questions ($P_{probe}$).}
\label{fig:prompt_probe}
\end{minipage}

\FloatBarrier
\section{AI Assistance Statement}

We used Gemini3 for polishing the writing to enhance clarity and readability.

\end{document}